\documentclass[preprint,12pt]{elsarticle}

\usepackage{amssymb}

\usepackage{booktabs}
\usepackage{graphicx}
\usepackage{multirow}
\usepackage{multicol}
\usepackage{amsmath}
\usepackage{makecell}
\usepackage{hyperref}
\usepackage{color}
\usepackage{wrapfig}

\newcommand{\etal}{\textit{et al.}}

\begin{document}

\begin{frontmatter}



\title{Weakly-supervised Pre-training for 3D Human Pose Estimation via Perspective Knowledge}



\author[]{Zhongwei Qiu}
\author[]{Kai Qiu}
\author[]{Jianlong Fu}
\author[]{Dongmei Fu}

\begin{abstract}
Modern deep learning-based 3D pose estimation approaches require plenty of 3D pose annotations. However, existing 3D datasets lack diversity, which limits the performance of current methods and their generalization ability. Although existing methods utilize 2D pose annotations to help 3D pose estimation, they mainly focus on extracting 2D structural constraints from 2D poses, ignoring the 3D information hidden in the images. In this paper, we propose a novel method to extract weak 3D information directly from 2D images without 3D pose supervision. Firstly, we utilize 2D pose annotations and perspective prior knowledge to generate the relationship of that keypoint is closer or farther from the camera, called relative depth. We collect a 2D pose dataset (MCPC) and generate relative depth labels. Based on MCPC, we propose a weakly-supervised pre-training (WSP) strategy to distinguish the depth relationship between two points in an image. WSP enables the learning of the relative depth of two keypoints on lots of in-the-wild images, which is more capable of predicting depth and generalization ability for 3D human pose estimation. After fine-tuning on 3D pose datasets, WSP achieves state-of-the-art results on two widely-used benchmarks.
\end{abstract}




\end{frontmatter}


\section{Introduction}

The goal of 3D human pose estimation is to estimate the 3D coordinates of human joints from 2D image, which provides accurate and fine-grained posture of the human body in the real 3D world. As an essential task in computer vision, 3D human pose estimation can be applied in human behavior understanding \cite{avola20192,mazzia2022action}, human-object interaction detection~\cite{li2020detailed}, and athletic training assistance~\cite{wang2019ai} and so on. Recently, with the breakthrough of deep convolutional neural network (CNN) \cite{he2016deep,xiao2018simple} and the disclosure of some large-scale datasets \cite{ionescu2013human3,mehta2017monocular,mehta2018single} for human pose estimation, many 3D human pose estimation methods \cite{sun2018integral,moon2019camera,qiu2019cross,kocabas2019self,iskakov2019learnable} have achieved remarkable improvements. However, the challenges of lacking rich 3D pose data and improving generalization ability still remain.

Monocular 3D human pose estimation heavily relies on large amounts of images and relative annotations of 3D human pose. At present, most 3D datasets of human pose are collected by a multi-view motion capture system (MoCap), which requires fine adjustment of equipment parameters and actors performing action performances in constraint places. Thus, collecting and labeling 3D human pose data consumes a lot of costs. Meanwhile, due to the limitation of the MoCap system, existing 3D pose estimation datasets \cite{ionescu2013human3,mehta2017monocular,mehta2018single} are collected in indoor lab or outdoor restricted space with several actors, which lack the diversity of background and human poses. A well-known issue \cite{zeng2020srnet} is that insufficient diversity of the training datasets will seriously affect the performance of 3D pose estimation network. The models trained on these constraint images usually fail to generalize in the real wild scenes, which limits the application of 3D human pose estimation models in the industry.

\begin{figure}
    \centering
    \includegraphics[width=\textwidth]{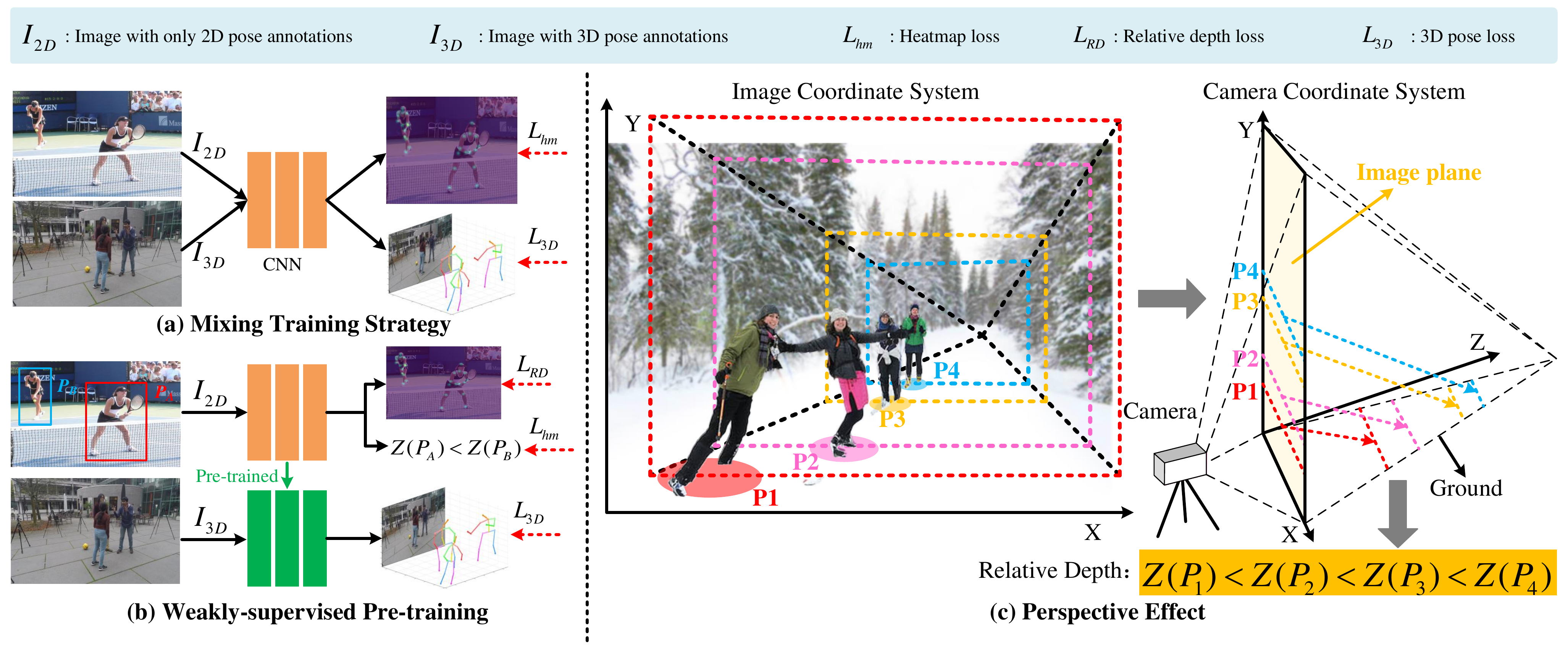}
    \caption{(a) Existing mixing training strategy (MTS)~\cite{sun2018integral,moon2019camera} train network by mixing images with and without 3D pose annotations. (b) Our weakly-supervised pre-training (WSP) approach train network on images without 3D pose annotation in a weakly-supervised way, with the supervision of relative depth. Then, we finetune network on images with 3D pose annotations. (c) The process of generating relative depths via perspective effect.}
    \label{fig:f1}
\end{figure}

To solve the above problems, 
many works \cite{zhou2017towards,sun2018integral,kocabas2019self,moon2019camera,iskakov2019learnable} train 3D pose models with the helping of some 2D pose datasets (e.g. MPII~\cite{andriluka20142d}, COCO~\cite{lin2014microsoft}). 
These mixing training strategy (MTS) can be summarized as Figure \ref{fig:f1} (a). They use 2D heatmaps as the supervision of the network when inputting images without 3D annotation, while 3D pose as the supervision when inputting images with 3D annotation. 
Benefited from over ten thousand images provided by these 2D datasets~\cite{andriluka20142d,lin2014microsoft}, annotated with accurate 2D pose in the natural scenes, they make the breakthrough of 3D human pose estimation. 
However, they just use the 2D pose as the supervision when training 3D pose models since no 3D annotations in these 2D datasets, ignoring the 3D information hidden in 2D images. 
For example, as shown in Figure \ref{fig:f1} (b), human can distinguish the relations of person $P_A$ is closer from the camera than person $P_B$. This weak 3D information can not be captured by just giving the supervision of 2D pose from single person. Other works have explored to alleviate the above problems by composting more training images~\cite{pishchulin2012articulated,chen2016synthesizing,varol2017learning,mehta2017monocular} or weakly supervised methods~\cite{kocabas2019self,chen2019weakly,wandt2019repnet} by using multi-view and projecting information. However, the composited images remain gap in realism with natural images, resulting in different data distribution between training data and testing data. Besides, camera parameters and multi-view images can not be provided in existing 2D datasets, which limits the application of these weakly supervised methods.

In this paper, we tackle the two problems:
1) the problems of lacking enough 3D human pose annotation, 
2) improving the generalization ability of 3D pose estimation model. 
Based on the observations of existing 2D human pose datasets, these real images can provide rich information of different background, diverse human poses, and different scales of human instances, which could help 3D human pose estimation. Moreover, we found that human could extract some weak 3D information directly from monocular 2D images by perspective knowledge. According to perspective 
effect, for an image captured by camera at a horizontal angle on the ground, the objects closer from camera are bigger than the objects farther from the camera. Meanwhile, for the points on the ground in this image, the points closer from camera have bigger $Y$-axis values in image coordinate system, which is named as relative depth. 

As shown in Figure \ref{fig:f1} (c), given a 2D image with multiple persons, it's easy for human to distinguish which person is closer from the camera. They make the verdict based on the empirical knowledge of foreshortening effects. Therefore, we explore to solve the above problems by utilizing large scale 2D human pose data and perspective knowledge. Additional 3D information and rich posture data in natural scenes, provided from 2D datasets, can improve the accuracy of 3D pose estimation and the generalization ability of 3D pose estimation model.
Different from previous methods \cite{zhou2017towards,sun2018integral,pavlakos2018ordinal,kocabas2019self,moon2019camera,yang20223d}, we extract weak 3D information directly from 2D images without 3D annotations or other extra manual labeling, which can significantly improve the performance of predicting 3D pose without extra costs of data collection and labeling. 

In order to learn relative depth of keypoints from 2D images, we collect a lager scale 2D dataset of human pose (MCPC) and generate relative depth according to the perspective knowledges. Since existing 2D dataset (MPII~\cite{andriluka20142d}, COCO~\cite{lin2014microsoft}, PoseTrack~\cite{andriluka2018posetrack} and CrowdPose~\cite{li2019crowdpose}) don't meet the requirements that multiple persons are in an image and each person has the annotations of ankle, we rearranged the images from the 2D datasets and labeled the annotations of relative depths of human bodies for each image by the perspective prior knowledge. Then, we propose a weakly-supervised pre-training (WSP) approach to learn weak 3D representation from MCPC dataset. 
WSP include two steps: pre-training on MCPC dataset and finetuning on 3D pose datasets. For pre-training, WSP firstly learns 2D heatmaps, which is used as spatial mask to extract keypoints feature. Then, WSP learns the relative depth of two keypoints from the keypoints features. For 3D human pose estimation, we finetune WSP on the 3D human pose estimation dataset. Since WSP learned weak 3D representation from a lot of real 2D images in the wild scenes, it can greatly improve the accuracy of 3D human pose estimation, especially in the depth prediction of 3D pose. In addition, WSP has strong generalization ability because it's pre-trained on a lot of real 2D images with diverse human poses and background in the wild scenes.

Our main contributions can be summarized as follows:
\begin{itemize}
    \item To the best of our knowledge, we are the first one, who uses the perspective knowledge to generate relative depths of keypoints for the pre-training of 3D human pose estimation. It's less costly since no need of 3D pose annotation or artificial relative depth labeling.
    \item We propose a large-scale dataset (MCPC), with over 53k images and 220k instances, with the annotations of relative depth. These data could help relieve the need for large amounts of 3D pose annotations and improve the generalization ability of the 3D pose model.
    \item We propose the WSP approach and pre-train it on MCPC dataset. After finetuning WSP on Human3.6M and MuCo-3DHP datasets, our approach outperforms previous approaches and achieve state-of-the-art results on these two benchmarks.

\end{itemize}

The remainder of this paper is organized as follows. We introduce the related work in Section \uppercase\expandafter{\romannumeral2}. Our approach and MCPC dataset is introduced in Section \uppercase\expandafter{\romannumeral3}. Experiments and analysis are presented in Section \uppercase\expandafter{\romannumeral4}. The Conclusions are summarized in Section \uppercase\expandafter{\romannumeral5}.

\section{Related Work}
We study the problems of 3D human pose estimation and try to improve the generalization ability of 3D pose estimation model. Here, we briefly review the related works on these problems.
\subsection{3D human pose estimation}
The progress in the field of 3D pose estimation in recent years benefit a lot from the development of deep learning.
These methods mainly fall into two paradigms, directly \textit{inferring} from images or \textit{lifting} from 2D poses.

The \textit{lifting} paradigm first conducts 2D human pose estimation on 2D images and then lifts 2D pose to 3D pose by a shallow network. Since there are a lot mature methods of 2D human pose estimation \cite{newell2016stacked,chen2018cascaded,xiao2018simple,sun2019deep,qiu2020dgcn,qiu2019learning,benzine2021single,sharma2021end}, many \textit{lifting} works \cite{martinez2017simple,cai2019exploiting,pavllo20193d,zhao2019semantic,zeng2020srnet} are mainly focus the \textit{lifting} stage. Martinez~\etal~\cite{martinez2017simple} proposes a simple fully connected network to lift the 3D pose from the 2D pose. Based on this work, some works study the influence of human pose structure by designing local or global network structures. Ci~\etal~\cite{ci2019optimizing} proposes a locally connected network to learn the local joint relations. Moreover, pose structure can be represented in a graph structure, with structural feature learning and 3D human pose estimation by graph convolutional network (GCN). Zhao~\etal~\cite{zhao2019semantic} proposes semantic GCN to capture the semantic information of human pose. These methods are lightweight and can be benefited from state-of-the-art 2D human pose models. However, the results of the \textit{lifting} methods awfully depend on the quality of the 2D pose. The learned network is over-fitting on datasets without seeing the real 2D images. Pavlakos~\etal~\cite{pavlakos2018ordinal} proposed to use a weaker supervision signal provided by the ordinal depths of human joints in the first stage. The estimated depths 2D coordinates are combined together as the input of $lifing$ network.

The \textit{inferring} methods directly regress 3D coordinates of human joints from the input 2D image. Sun~\etal~\cite{sun2018integral} proposes to learn 3D heatmaps by a backbone network (ResNet), then regressing 3D coordinates by integral regression on 3D heatmaps, which is the simplest and state-of-the-art approach. This is the mainstream framework of inferring-based 3D human pose estimation. Based on this framework, many works try to study self-supervised methods~\cite{kocabas2019self}, multi-view methods~\cite{qiu2019cross,iskakov2019learnable} and multi-person 3D human pose estimation~\cite{moon2019camera,qiu2022dynamic}. Compared with \textit{lifting} methods, the \textit{inferring} methods study extracting features from the 2D images in an end-to-end way without overfitting on 2D pose data in a special dataset, which has greater application potential.  

\subsection{The problem of lacking 3D pose annotation}
The success of existing 3D human pose estimation methods relies on multiple 3D datasets~\cite{mehta2017monocular,ionescu2013human3,mehta2018single} with a larger number of 3D pose annotations. However, most of the existing 3D datasets are collected in the indoor lab or restricted outdoor space with several actors, who perform specific action performances. Therefore, the performance of existing 3D human pose estimation methods is limited by insufficient data on the 3D pose. The mainstream ways to address this problem are adding 2D pose datasets, generating more training images, and weakly supervised learning. 

\subsubsection{Mixing 2D pose dataset training}
Some works~\cite{zhou2017towards,sun2018integral,habibie2019wild,yang20183d} have studied the mixed training of 2D and 3D human pose datasets. Zhou~\etal~\cite{zhou2017towards} a weakly-supervised transfer learning network with a 2D pose estimation sub-network and a 3D depth regression sub-network, which uses mixed 2D and 3D labels. Sun~\etal~\cite{sun2018integral} proposed a unified framework to mix 2D and 3D labels by integral regression on 3D heatmaps. Yang~\etal~\cite{yang20183d} proposed an adversarial learning framework, which distills the 3D human pose structures learned from the fully annotated dataset to in-the-wild images with only 2D pose annotations. Habibie~\etal~\cite{habibie2019wild} introduced a network architecture that comprises a new disentangled hidden space encoding of explicit 2D and 3D features and uses supervision by a new learned projection model from a predicted 3D pose. These methods have extensively verified that 2D labels can improve the accuracy of 3D pose estimation. However, these methods just extract 2D information from images and 2D labels to enhance the constraint of the 3D pose, which limits the role of 2D labels and images. In this paper, We directly extract 3D information from 2D labels and images to improve the performance of 3D human pose estimation.

\subsubsection{Data synthesis and data augmentation}
Some approaches try to generate more training images through data augmentation~\cite{varol2017learning,rogez2016mocap,pishchulin2012articulated,mehta2017monocular}. Varol~\etal~\cite{varol2017learning} propose to generate more training images by computer graphics rendering. Rogez~\etal~\cite{rogez2016mocap} propose mocap-guided data augmentation for 3D human pose estimation in the wild. Pishchulin~\etal~\cite{pishchulin2012articulated} fitted 3D models to humans and deformed them to generate new images in multiple views. Mehta~\etal~\cite{mehta2017monocular} tried to generate more images by changing the background and clothing in existing images. Although these methods improve the performance of the 3D human pose estimation model by generating more training images, the fundamental problem has not been solved since the gap between synthesized images and real images still exists.

\subsubsection{Weakly-supervised Learning}
Other works try to solve this problem by weakly supervised methods \cite{kocabas2019self,chen2019weakly,wandt2019repnet}. Kocabas~\etal~\cite{kocabas2019self} introduces a self-supervised learning method to train a 3D pose model by generating 3D pose labels according to multi-view information and camera parameters. Chen~\etal~\cite{chen2019weakly} proposes to learn a geometry representation from multi-view information. The learned representation could map to a 3D pose with a shallow network and less annotated training samples. Wandt~\etal~\cite{wandt2019repnet} proposes an adversarial reprojection network to train the 3D pose model. They predict 3D pose and camera parameters at the same time, then project the 3D pose to 2D pose by camera information. Although these methods reduce the reliance on 3D pose annotations, the performance gap between weakly supervised methods and fully supervised methods is huge. Meanwhile, camera parameters and multi-view formation can’t be provided in existing monocular images, which limits the application of weakly supervised methods.

\subsection{Generalization in the wild}
The above methods to solve the problems of lacking 3D pose data have alleviated the problems of generalization \cite{zeng2020srnet,fang2018learning,wang2020predicting}. Other methods study the generalization according to different viewpoints. Fang~\etal~\cite{fang2018learning} utilized virtual camera simulation to generate more 2D pose data from different viewpoints and designed a generative adversarial network (GAN) to learn pose grammar, which can encode the human body configuration for 3D human pose estimation. Wang~\etal~\cite{wang2020predicting} proposed a viewpoint prediction network to improve cross-dataset generalization for 3D human pose estimation. Zeng~\etal~\cite{zeng2020srnet} introduced a split-and-recombine approach to improve generalization by recombining rare or unseen poses in the training set. These methods try to improve the generalization by enhancing the adaptability of the model. However, they can't handle the real images in the wild since they are designed especially for existing datasets.

\section{Approach}

To address the problems of lacking 3D labels and generalizing the 3D pose model to the in-the-wild scenes, we propose a framework that pre-training the 3D pose network on the 2D dataset and finetuning this network on the downstream task of 3D human pose estimation. We propose an approach to directly extract relative depth information on RGB images via perspective knowledge, which is used as the supervision of learning 3D representation for weakly-supervised pre-training (WSP). The WSP strategy enables the network to learn how to distinguish the relative depth of two points in an image, while can help 3D pose estimation.
For pre-training, we collected a dataset called MCPC. We pre-trained WSP on MCPC and finetuned on 3D pose datasets.

\subsection{Generating Relative Depth via Perspective Knowledge}
\label{seq:rd}
As shown in Figure \ref{fig:f1} (c), for an image with multiple persons, it's easy for humans to estimate that person $P1$ is closer to the camera than the remaining people ($P2, P3, P4$). We can estimate the relative depth according to the perspective knowledge that objects close to the camera in the image are larger, while objects far away from the camera are smaller. Meanwhile, assuming that the 2D images are all captured by the camera at a horizontal angle on the ground, then a point on the ground in the image with a greater $Y$-axis value is closer to the camera than other points on the ground in the image with smaller $Y$-axis value. These are two of the experience of perspective knowledge. In the following, we introduce the details to generate relative depth.

For the joints (e.g., head, ankle, etc.) that belong to the persons in an image, the goal is to obtain their relative depths. Supposing that the human body is standing on the ground, we can simplify that ankle points are ground points. Then, we can obtain the relative depths of ankles belonging to different persons according to the above knowledge of human experience. For example, for ankle points $P_A^{~ankle}$ belong to person $A$ and ankle points $P_B^{~ankle}$ belong to person $B$ in Figure \ref{fig:f2} (a), the relative depths between $P_A^{~ankle}$ and $P_B^{~ankle}$ is 
\begin{equation}
\label{eq:rd_ankle}
    Z^C(P_A^{~ankle}) < Z^C(P_B^{~ankle})
\end{equation}
since 
\begin{equation}
    \label{eq:y_rd}
    Y^I(P_A^{~ankle}) > Y^I(P_B^{~ankle})~,
\end{equation}
where $Z^C(\cdot)$ means the depth in camera coordinate system. $Y^I(\cdot)$ means the $Y$-axis value in image coordinates system.

Now, the relative depth between person $A$ and person $B$ can be obtained according to Equation \ref{eq:rd_ankle}, that is 
\begin{equation}
\label{eq:rd_ab}
    Z^C(A) < Z^C(B)~,
\end{equation}
which means that person $A$ is closer to the camera than person $B$. Although the all joints on person $A$ are not always closer to the camera than the all joints on person $B$, to obtain the relative of all human joints, let's simplify the condition that 
\begin{equation}
\label{eq:rd_all_kps}
    Z^C(P_A^{j}) < Z^C(P_B^{j}),~j=1,...,J~,
\end{equation}
where $J$ is the number of human joints. Equation \ref{eq:rd_all_kps} means that the joints on person $A$ are closer to the camera than the joints on person $B$. 

\textbf{Note} that there are some negative examples when person $A$ is extremely close to person $B$, such as $A$ hugs $B$. Although we ignore these cases here, our experimental results show that our model has the capability to handle these incorrect labels, which can be regarded as noises.

We obtain relative depths of human joints on 2D datasets according to Equation \ref{eq:rd_ankle} - \ref{eq:rd_all_kps}, which are used as the supervision to train network. 

\subsection{MCPC Dataset}
To tackle the problem of lacking enough 3D pose data with diversified posture and background, we want to extract weak 3D information from 2D datasets for 3D human pose estimation. There are several 2D datasets with a lot of 2D pose annotations, which are MPII~\cite{andriluka20142d}, COCO~\cite{lin2014microsoft}, Posetrack~\cite{andriluka2018posetrack} and CrowdPose~\cite{li2019crowdpose}. However, some images in these datasets may only contain a single person or multiple persons without the annotations of the ankle. This can't meet the requirements to generate relative depth. Therefore, we select images from these 2D datasets to form a new dataset, named MCPC.

The images of MCPC come from four 2D datasets: MPII, COCO, Posetrack, and CrowdPose. We collect images according to the two principles: 1)There are at least two persons in an image, 2)The ankle joints must be annotated in these persons.
After filtering images, we collect over 53k images with more than 220k human instances. Since each image in MCPC has different human subjects with different postures and backgrounds, MCPC is a larger scale 2D dataset in natural scenes. For each image in the MCPC dataset, we generate relative depths for each human joint according to the perspective knowledge introduced in Section \ref{seq:rd}.

We make a comparison of several widely-used 2D datasets and 3D datasets in Table \ref{table:mcpc}. Existing 3D pose datasets (e.g. Human3.6M, MuCo), are collected in indoor environments. Although there are over 300k images in these datasets, there are only 7 and 8 different human subjects in the Human3.6M and MuCo datasets, respectively. The images in these datasets have a high degree of similarity since they are captured with several subjects in the same place. The images in 2D datasets (e.g. MPII, COCO, Posetrack, and CrowdPose) are captured in different environments with different human subjects. However, they have no 3D labels. The MCPC dataset has multiple instances in an image with 2D pose labels and relative depth labels. The relative depth labels are the key information to improve the performance of 3D human pose estimation and the generalization ability of the 3D pose model.

\begin{table}
    \centering
  \renewcommand\tabcolsep{8pt}
  \caption{Comparison of 2D and 3D pose estimation datasets. ``In." means indoor and ``Out." means outdoor. ``RD" means relative depth labels. Existing 3D pose estimation datasets (Human3.6M~\cite{ionescu2013human3}, Muco~\cite{mehta2018single}) are collected with a few subjects performing different actions in limited environments.}
  \vspace{0.2cm}
\resizebox{\linewidth}{!}{
  \begin{tabular}{lccccc}
    
    \hline
    
    \hline
    Dataset & Images & Instances & Subjects & Environment & Labels\\
    \hline
    Human3.6M~\cite{ionescu2013human3} & $\sim$312k  & $\sim$312k & 7 & In. & 3D pose \\
    MuCo~\cite{mehta2018single} & $\sim$400k  & $\sim$400k & 8 & In. & 3D pose \\
    \hline
    MPII~\cite{andriluka20142d} & $\sim$17k  & $\sim$28k & $\sim$28k & In./Out. & 2D pose \\
    COCO~\cite{lin2014microsoft} & $\sim$14k  & $\sim$56k & $\sim$56k & In./Out. & 2D pose \\
    Posetrack~\cite{andriluka2018posetrack} & $\sim$12k  & $\sim$56k & $\sim$56k & In./Out. & 2D pose \\
    CrowdPose~\cite{li2019crowdpose} & $\sim$20k  & $\sim$80k & $\sim$80k & In./Out. & 2D pose \\
    \hline
    MCPC (Ours) & $\sim$53k  & $\sim$220k & $\sim$220k & In./Out. & \makecell[c]{2D pose+RD}  \\
    
    \hline
    
    \hline
  \end{tabular}
  }
  \label{table:mcpc}
\end{table}

\subsection{Framework}
\begin{figure}
    \centering
    \includegraphics[width=\textwidth]{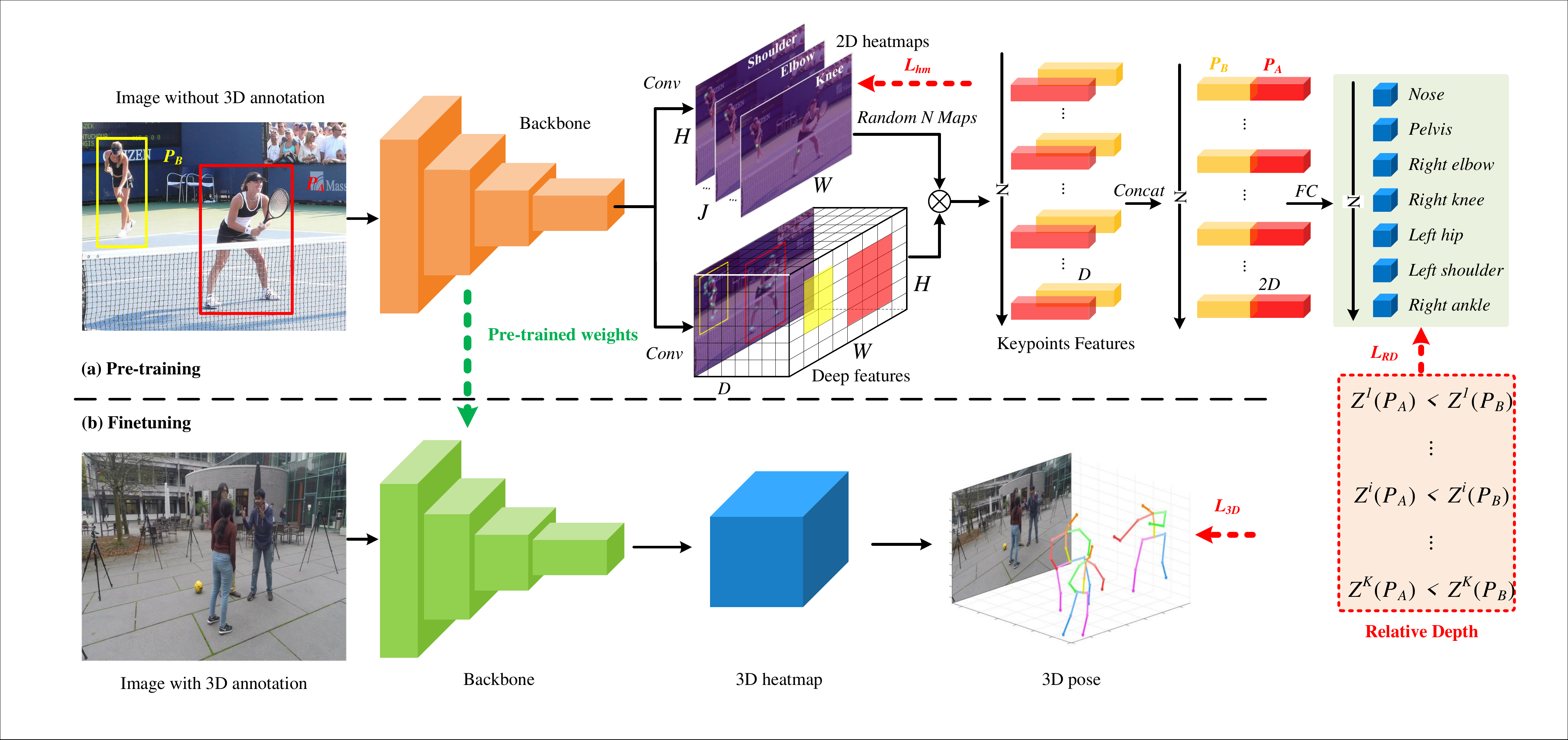}
    \caption{The WSP framework includes (a) pre-training on MCPC dataset without 3D pose annotations and (b) finetuning on 3D pose estimation dataset. $\otimes$ denotes element-wise multiplication. $FC$ means fully connection layers. The pre-training learns better 3D representation by learning relative depth, which is beneficial for 3D pose estimation.
    }
    \label{fig:f2}
\end{figure}

The goal of monocular 3D human pose estimation is to recover the absolute camera-centered coordinates of human joints from a 2D image. The overall architecture of our proposed framework is shown in Figure \ref{fig:f2}, which includes two steps: pre-training on 2D datasets and finetuning on 3D datasets.

Firstly, we design a network WSP to extract the keypoints features. Given an RGB image $I_{2D}$~, two-person pairs are generated by randomly selecting from multiple person instances in the image. Then, the image patches cropped from the whole image according to the bounding box of two-person pairs are sent into WSP. Inputting an image patch $I_{AB}$, which contains person instance $A$ and $B$. The outputs of WSP are the relative depths between human joints $P_A$ in person $A$ and human joints $P_B$ in person $B$. If person $A$ is closer to the camera than person $B$, the relative depths can be denoted as $Z(P_A^j) < Z(P_B^j), j\in[1,J]$, in which $J$ is the number of human joints. 
Note that there may be more than two persons in an image patch because of the large pose scales of person $A$ and person $B$. The other persons in this image patch $I_{AB}$ are ignored, which means WSP just outputs the relative depths of person $A$ and person $B$. The details of WSP and how to train WSP on 2D dataset are introduced in Section \ref{section_krdln}.

After pre-training WSP on 2D datasets, WSP has the ability to distinguish which joint is closer to the camera than other joints. Then, WSP is finetuned on 3D human pose estimation datasets. The pre-trained WSP can accelerate the convergence of the model for 3D human pose estimation since it has learned weak 3D representation. The pre-trained WSP also provides a good starting point to train the 3D pose model and improves the generalization ability of the model.

\subsection{Relative Depth Pre-training Network}
\label{section_krdln}
\subsubsection{Model Design}
WSP estimates the relative depths between human joints from a cropped image with two-person pair, which is shown in Figure \ref{fig:f3}. WSP consists of three parts: backbone, relative depth head, and 2D keypoints heatmap head.

The first one, backbone, extracts deep features from the input image by using ResNet. Based on the extracted deep features, the second part, the 2D keypoints heatmap head predict the keypoints heatmap by 3 deconvolution layers. Different from other methods to obtain 2D coordinates from 2D heatmaps, the estimated heatmaps here are used as masks to filter deep features for special human joints. 

The relative depth head estimates the relative depths of human joints when given the deep features of two points. The inputs of relative depth head are deep features filtered by keypoints heatmaps. Deep features $F_d$ has a size of (D, H, W), and 2D heatmap $M$ has a size of (J, H, W). D is the channel number. H and W are the height and width of heatmaps. J represents J joints. The keypoints deep features are extracted as follows
\begin{equation}
    \label{eq:matmul}
    F^j = F_d \otimes repeat(M^j, D), j=1,...,J
\end{equation}
where $j$ denotes $j$-th joint. $repeat$ means the operation of repeating $M^j$ of size $H\times W$ into size $D\times H \times W$. $\otimes$ denotes element-wise multiplication. We ignore other persons who may are cropped in this image patch and their relative depths are estimated in other image patches. $M^j$ represents the 2D heatmaps for the $j$-th keypoint. $F^j$ represents the filtered deep features for the $j$-th keypoint. Then, for each person, $F^j$ is extracted into a vector of size $D$ by the pooling layer within the bounding box of this person. Thus, the extracted deep features have a size of (J, D).

Then, we randomly select $N~(N<J)$ keypoints for each person to estimate their relative depths. We use a fully connected network to classify the relative depths of selected keypoints. As shown in Figure \ref{fig:f2}, for the person $A$ and person $B$ in the input image, we firstly obtain the filtered deep feature $F(P_A)$ and $F(P_B)$ by Equation \ref{eq:matmul}, which have a size of (J, D). After randomly selecting, we get $F_s(P_A)$ and $F_s(P_B)$, which have a size of (N, D). For each randomly selected keypoint, we concatenated $F_s^j(P_A)$ and $F_s^j(P_B)$ and send them into the fully connected network to estimate their relative depths. The predicted relative depths can be denoted as 
\begin{equation}
    \label{eq:rd_head}
    R_{AB}^j = f(F_s(P_A^j)\oplus F_s(P_B^j))~,
\end{equation}
where $\oplus$ represents concatenate operation. $f(\cdot)$ represents fully connected network of input dimension $2\times D$ and output dimension $1$. $R_{AB}^j$ represents the estimated relative depths for $j$-th joint in person $A$ and person $B$.

\subsubsection{Loss Function}
Given an image patch $I_{AB}$ cropped from original image $I_{2D}$, WSP estimates the relative depths $R_{AB}^j$ for each joints in person $A$ and $B$ according to the Equation \ref{eq:matmul} and \ref{eq:rd_head}. The ground-truth of relative depths is $\hat{R}_{AB}^j$, which can be generated according to Equation \ref{eq:rd_ankle}, \ref{eq:y_rd}, \ref{eq:rd_ab} and \ref{eq:rd_all_kps}. To numerical represent relative depths, we define that 
\begin{eqnarray}
\hat{R}_{AB}^j=\begin{cases}
1,\quad Z(P_A^j) \leq Z(P_B^j) \\
0,\quad Z(P_A^j) > Z(P_B^j)
\end{cases}
\end{eqnarray}
where $P_A^j$, $P_B^j$ and $Z(\cdot)$ are the $j$-th joint in person $A$, the $j$-th joint in person $B$ and the operation to get depths, respectively.

We use cross-entropy loss and $L_2$ loss for relative depth learning and heatmap learning, respectively. The total training loss is 
\begin{equation}
\label{eq:loss_krdln}
    \mathcal{L} = \mathcal{L}_{hm} + \alpha \times \mathcal{L}_{rd}~,
\end{equation}

where $\mathcal{L}_{hm}$, $\mathcal{L}_{rd}$ and $\alpha$ are the loss for keypoints heatmap, keypoints relative depth, and loss weight, respectively. 

$\mathcal{L}_{rd}$ is cross-entropy loss, and
\begin{equation}
    \mathcal{L}_{rd} = \frac{1}{J} \sum_j^J -[\hat{R}^j log(R^j) + (1 - \hat{R}^j) log(1 - R^j)]~,
\end{equation}
where $J$, $\hat{R}^j$, and $R^j$ are the number of human joints, the ground-truth of relative depths for $j$-th joint, and the relative depth estimated by WSP for $j$-th joint, respectively.

$\mathcal{L}_{hm}$ is $L_2$ loss, and
\begin{equation}
    \mathcal{L}_{hm} = \frac{1}{JK} \sum_k^K \sum_j^J ||M_k^j - \hat{M}_k^j||_2~,
\end{equation}
where $J$, $K$, $M_k^j$ and $\hat{M}_k^j$ are the number of human joints, the number of estimated persons in the input image, the predicted keypoints heatmaps, and the ground-truth of keypoints heatmaps for $j$-th joint in person $k$, respectively.

\subsection{Finetuning WSP}
\subsubsection{Model design}
We finetune WSP on the 3D human pose estimation dataset. For 3D human pose estimation, as \cite{sun2018integral,kocabas2019self,moon2019camera}, we add a head to learn 3D keypoints heatmap. The 3D coordinates can be obtained by integrating 3D heatmaps. The integrating operation to get $x$ values of 3D coordinates can be formulated as 
\begin{equation}
\label{eq:integeate}
    C_{x}^j = \sum_{x=1}^W \sum_{z=1}^D \sum_{y=1}^H \hat{M}^j(p)~,
\end{equation}
where $p$ represents the points in 3D heatmap $\hat{M}$. $C_{x}^j$ is the $x$-axis values of 3D coordinates for $j$-th joints. $C_{y}^j$ and $C_{z}^j$ are obtained also according to the Equation \ref{eq:integeate} by changing the order of integrating.

\subsubsection{Loss Function}
Following the suggestion from \cite{sun2018integral}, we use $L1$ loss to optimize the errors of coordinates. The training loss of finetuning WSP is 
\begin{equation}
    \mathcal{L}_{3d} = \frac{1}{J}\sum_j^J||C^j - \hat{C}^j||_1~,
\end{equation}
where $J$, $C^j$ and $\hat{C}^j$ are the numbers of human joints, the estimated coordinates by our network and the ground-truth coordinates, respectively.

\section{Experiments}
\subsection{Dataset and Evaluation Metrics}
Our WSP is pre-trained on MCPC dataset, then finetuned on Human3.6M and MuCo-3DHP datasets. Here, wo introduce the experimental protocol and evaluation metrics.

\subsubsection{MCPC Dataset}
The images in MCPC dataset for training are all collected from the train set in MPII, COCO, PoseTrack, and CrowdPose dataset. The train set of MCPC contains over 53k images with more than 220k persons. The images in the test set of MCPC are all from the test set in these four 2D datasets. The test set of MCPC contains over 11k images with more than 18k two-person pairs.

For relative depth pre-training, we evaluate the predicted relative depths of human joints on the test set of the MCPC dataset. Since it is formulated as a classification task, we use AUC, Accuracy, Precision, Recall, and F1 score to evaluate the predicted relative depths. The threshold for accuracy, precision, recall, and F1 score is 0.5.

\subsubsection{Human3.6M Dataset}
Human3.6M~\cite{ionescu2013human3} is the largest and most widely-used benchmark for 3D human pose estimation. This dataset is captured by the MoCap system in the lab environment. It consists of 3.6 million videos frames from 4 cameras' viewpoints. 11 subjects perform 15 activities, such as discussion, eating, etc. The appearance of the human body and the background of the images is simple since the limited subjects and lab environment. Accurate 3D annotations of the human pose are generated by a motion capture system in each frame.

For evaluation, we use mean per joint position error (MPJPE), which is widely used in previous works. Some works also use another widely-used metric, PA MPJPE. They firstly align the predicted 3D pose and ground truth 3D pose with a rigid transformation using Procrustes Analysis and then compute MPJPE, which is called PA MPJPE.

\subsubsection{MuCo-3DHP and MuPoTS-3D Datasets}
MuCo-3DHP and MuPoTS-3D Datasets are 3D multi-person pose estimation datasets proposed by Mehta~\etal~\cite{mehta2018single}. MuCo-3DHP is the training set, which is generated by compositing the existing MPI-INF-3DHP 3D single-person pose estimation dataset~\cite{mehta2017monocular}. The testing set, MuPoTS-3D dataset is captured outdoors and includes 20 real-world scenes with ground-truth 3D poses for up to three subjects. The 3D annotations of the MuPoTS-3D dataset are obtained with a multi-view marker-less motion capture system.

For evaluation, 3DPCK, a 3D percentage of correct keypoints is used after alignment with ground truth. It treats a joint’s prediction as correct if it lies within 15cm from the ground-truth joint location. For more rigorous testing, we also additionally use 3DPCK with thresholds of 10cm, 11cm, 12cm, 13cm, and 14cm, respectively.

\begin{table}
    \centering
  \renewcommand\tabcolsep{20pt}
  \caption{Ablations on MCPC dataset in \textit{\textbf{RD Protocol 1}}.}
\vspace{0.2cm}
\resizebox{\columnwidth}{!}{
  \begin{tabular}{lcccccc}
 
    \hline
    
    \hline
    Settings & $\Delta S$ & AUC(\%) & Acc(\%) & P(\%) & R(\%) & F1(\%)\\
    \hline
    WSP-A & $\leq$1.0 & 95.39 & 88.03 & 88.04 & 88.04 & 88.04\\
    WSP-B & $\leq$0.5 & 94.89 & 87.29 & 87.29 & 87.30 & 87.29\\
    WSP-C & $\leq$0.3 & 92.78 & 84.62 & 84.67 & 84.59 & 84.63\\
    WSP-D & $\leq$0.1 & 90.34 & 81.66 & 81.69 & 81.62 & 81.65\\
    \hline
    
    \hline
  \end{tabular}
  }
   \label{table:rdl_scale_eval}
\end{table}

\begin{table}
    \centering
  \small
  \renewcommand\tabcolsep{15pt}
  \caption{Ablations on MCPC dataset in \textit{\textbf{RD Protocol 2}}. RA means remove the ankle parts in image for testing.}
\vspace{0.2cm}
\resizebox{\columnwidth}{!}{
  \begin{tabular}{lccccccc}
    \hline
    
    \hline
    Settings & RA & $\Delta S$ & AUC(\%) & Acc(\%) & P(\%) & R(\%) & F1(\%)\\
    \hline
    WSP-A & $\times $  & $\leq$1.0 & 95.39 & 88.03 & 88.04 & 88.04 & 88.04\\
    WSP-E & $\surd$    & $\leq$1.0 & 93.72 & 85.70 & 85.66 & 85.70 & 85.68\\
    WSP-F & $\surd$    & $\leq$0.5 & 92.92 & 84.63 & 84.64 & 84.63 & 84.63\\
    WSP-G & $\surd$    & $\leq$0.3 & 90.21 & 81.56 & 81.50 & 81.67 & 81.58\\
    WSP-H & $\surd$    & $\leq$0.1 & 87.89 & 79.00 & 78.96 & 79.03 & 79.00\\
    \hline
    
    \hline
  \end{tabular}
  }
  \label{table:rdl_crop_eval}
\end{table}

\subsection{Experimental Protocol}
\subsubsection{MCPC Dataset}
We formulate the task of estimating relative depths as distinguishing which human joint is closer to the camera. When training, for each image in mini-batch, we randomly select two persons in this image to estimate their relative depths of human joints. Due to the different scales of the two-person pairs, we define that
\begin{equation}
\label{eq:scale}
    S = ||P_{ht} - P_{hd}||_2~,
\end{equation}
where $S$ denotes the scale of the person. $P_{ht}$ and $P_{hd}$ are the head\_top joint and head\_down joint in the person, respectively.

For person $A$ and person $B$, the normalized scale gap ratio $\Delta S$ is denoted as 
\begin{equation}
\label{eq:scale_ratio}
    \Delta S = \frac{|S_A - S_B|}{max(S_A,~S_B)}~,
\end{equation}
where $S_A$, $S_B$ are the scales of person $A$ and person $B$ generated by Equation \ref{eq:scale}. $0 \leq \Delta S \leq 1$.

In order to fully verify that WSP has learned the knowledge of estimating relative depths, we define two protocols. \textit{\textbf{RD protocol 1}}: we evaluate the estimated relative depths in different scale gap ratios ($\Delta S = 1.0, 0.5, 0.3, 0.1$). The smaller $\Delta S$ indicates that the person $A$ and person $B$ have similar scales, which means it's more difficult to estimate their relative depths.
\textit{\textbf{RD Protocol 2}}: Removing the human ankle parts in the input image for testing. \textit{\textbf{RD Protocol 2}} is designed to study if WSP still can distinguish the relative depth without the ankle clues.

\subsubsection{Human3.6M Dataset}
Three experimental protocols are used. Following previous works~\cite{moon2019camera,sun2018integral}, the first one uses five subjects (S1, S5, S6, S7, S8) as training data and the other two subjects (S9, S11) as testing data. We denote it as the \textit{\textbf{Subject Protocol 1}}, which uses MPJPE as the evaluation metric. The second one uses six subjects (S1, S5, S6, S7, S8, S9) as the training data and the other one subject (S11) as testing data. We denote it as the \textit{\textbf{Subject Protocol 2}}, which uses PA MPJPE as the evaluation metric. To evaluate the generalization of 3D pose estimation model, we also use another widely-used protocol following \cite{martinez2017simple,zeng2020srnet}, called \textit{\textbf{Cross Action Protocol}}, which uses MPJPE as the evaluation metric. \textit{\textbf{Cross Action Protocol}} trains on only one of the 15 actions in the Human3.6M dataset and tests on all actions, which is designed to test the generalization ability of the 3D pose model.

\subsubsection{MuCo-3DHP and MuPoTS-3D Datasets}
Following the previous work~\cite{moon2019camera}, we use 400k frames composited from the MuCo-3DHP dataset for training, in which half of the images are background augmented. As the setting of previous works, we also use the COCO dataset for augmentation. When training models, each mini-batch consists of half MuCo-3DHP  and half COCO images. We use the strategy of auxiliary training as \cite{sun2018integral,moon2019camera} since the COCO dataset has no annotations of the 3D pose.

\begin{table}
    \centering
  \renewcommand\tabcolsep{6pt}
  \caption{Comparison with baseline with no relative depth pre-training. For fair comparison, PoseNet and WSP are finetuned on Human3.6M (HM36) dataset and finetuned on mixed HM36 and MCPC dataset.}
  \vspace{0.2cm}
\resizebox{\columnwidth}{!}{
  \begin{tabular}{c|c|c|c|c|c|c|c|c|c|c|c|c}
    \hline
    
    \hline
    \multirow{2}{*}{Methods} & \multirow{2}{*}{Backbone} &\multirow{2}{*}{Dataset}& \multicolumn{10}{c}{MPJPE(mm)} \\
    \cline{4-13}
    &&& Action Avg &$\Delta$ & Joints Avg &$\Delta$ & X &$\Delta$ & Y &$\Delta$ & Z&$\Delta$ \\
    \hline
    PoseNet~\cite{moon2019camera} & \multirow{2}{*}{R-50} & \multirow{2}{*}{HM36} & 67.5 &\multirow{2}{*}{\color{red}{$\downarrow$13.2}} & 59.9 & \multirow{2}{*}{\color{red}{$\downarrow$12.3}}& 22.0 &\multirow{2}{*}{$\downarrow$4.6} & 24.1 & \multirow{2}{*}{$\downarrow$5.4}& 50.2 &\multirow{2}{*}{\color{red}{$\downarrow$10}}\\
    \textbf{WSP(Ours)} &  &  & 54.3 & & 47.6 & & 17.4 & & 18.7 & & 40.2 & \\
    
    \hline
    PoseNet~\cite{moon2019camera} & \multirow{2}{*}{R-50} & \multirow{2}{*}{HM36+MCPC} & 52.9 &\multirow{2}{*}{\color{red}{$\downarrow$5.4}} & 47.7 & \multirow{2}{*}{\color{red}{$\downarrow$5.4}}& 18.8 &\multirow{2}{*}{$\downarrow$2.5} & 18.3 & \multirow{2}{*}{$\downarrow$2.2}& 39.9 &\multirow{2}{*}{\color{red}{$\downarrow$4.4}}\\
    
    \textbf{WSP(Ours)} &  &  & 47.5 & & 42.3 & & 16.3 & & 16.1 & & 35.5 & \\
    
\hline

\hline
  \end{tabular}
  }
  \label{table:finetune_hm36}
\end{table}

\subsection{Implementation Details}
\subsubsection{Relative Depth Pre-training}
We use ResNet pre-trained on ImageNet as the backbone of WSP. Then, we train WSP on the MCPC dataset. We summarize the hyper-parameters as follows for pre-training: the loss weight $\alpha$ (in Equation \ref{eq:loss_krdln}) equals 50. The input image size is $256\times 256$ and the heatmap size is $64\times 64$. The batch size is 64 and the initiating learning rate is 0.001.
The learning rate is decay to 0.0001 and 0.00001 at $30$th epoch and $40$th epoch, respectively. The total epoch number for training is 50. We conduct the experiments using PyTorch on the Linux platform, and 2 V100 GPUs are used. It cost about 50 hours to train WSP on the MCPC dataset.

\subsubsection{Finetuning on 3D Human Pose Estimation Dataset}
We use WSP pre-trained on the MCPC dataset as the backbone network for 3D human pose estimation. We remove the original heads of WSP and add an up-sampling head to estimate the 3D human pose. We finetune the network on Human3.6M and MuCo-3DHP datasets, respectively. We summarize the hyper-parameters as follows for WSP finetuning: The input image size is $256\times 256$ and the heatmap size is $64\times 64$. The batch size is 64 and the initiating learning rate is 0.001.
The learning rate is decay to 0.0001 and 0.00001 at $17$th epoch and $21$th epoch, respectively. The total epoch numbers for training is 25. We conduct the experiments using PyTorch on the Linux platform, and 2 V100 GPUs are used. It cost about 8 hours and 20 hours to finetune WSP on Human3.6M and MuCo-3DHP datasets, respectively. 

\begin{table}
  \caption{Comparison with State-of-the-art Methods on the 3D pose dataset (Human3.6M) in \textit{\textbf{Subject Protocol 1}} with MPJPE and \textit{\textbf{Subject Protocol 2}} with PA MPJPE. Bold
indicates the best and \underline{underline} indicates the second best}
  \vspace{0.2cm}

\renewcommand\tabcolsep{3pt}
  \resizebox{\textwidth}{!}{
  \begin{tabular}{lccccccccccccccccc}
    \hline
    
    \hline
    MPJPE(mm) & & Dir. & Dis. & Eat. & Gre. & Phoe. & Pho. & Pos. & Pur. & Sit. & SiD. & Smo. & Wai. & WaD. & Wal. & WaT. & Avg\\
    \hline
    Jahangiri~\etal~\cite{jahangiri2017generating}& ICCV'17& 74.4 & 66.7 & 67.9 & 75.2 & 77.3 & 70.6 & 64.5 & 95.6 & 127.3 & 79.6 & 79.1 & 73.4 & 67.4 & 71.8 & 72.8 & 77.6\\
    Mehta~\etal~\cite{mehta2017monocular}& 3DV'17 & 57.5 & 68.6 & 59.6 & 67.3 & 78.1 & 56.9 & 69.1 & 98.0 & 117.5 & 69.5 & 82.4 & 68.0 & 55.3 & 76.5 & 61.4 & 72.9\\
    Martinez~\etal~\cite{martinez2017simple}& ICCV'17& 51.8 & 56.2 & 58.1 & 59.0 & 69.5 & 55.2 & 58.1 & 74.0 & 94.6 & 62.3 & 78.4 & 59.1 & 49.5 & 65.1 & 52.4 & 62.9\\
    Fang~\etal~\cite{fang2018learning}& AAAI'18 & 50.1 & 54.3 & 57.0 & 57.1 & 66.6 & 53.4 & 55.7 & 72.8 & 88.6 & 60.3 & 73.3 & 57.7 & 47.5 & 62.7 & 50.6 & 60.4\\
    Sun~\etal~\cite{sun2017compositional}& ECCV'17  & 52.8 & 54.8 & 54.2 & 54.3 & 61.8 & 53.1 & 53.6 & 71.7 & 86.7 & 61.5 & 67.2 & 53.4 & 47.1 & 61.6 & 63.4 & 59.1\\
    Pavlakos~\etal~\cite{pavlakos2018ordinal}& CVPR'18 & 48.5 & 54.4 & 54.4 & 52.0 & 59.4 & 65.3 & 49.9 & 52.9 & 65.8 & 71.1 & 56.6 & 52.9 & 60.9 & 44.7 & 47.8 & 56.2\\
    Sun~\etal~\cite{sun2018integral}& ECCV'18  & 47.5 & 47.7 & 49.5 & 50.2 & 51.4 & \textbf{43.8} & 46.4 & 58.9 & 65.7 &\textbf{49.4} & 55.8 & 47.8 & \textbf{38.9} & 49.0 & 43.8 & 49.6\\
    Moon~\etal~\cite{moon2019camera}& ICCV'19 & 50.5 & 55.7 & 50.1 & 51.7 & 53.9 & 46.8 & 50.0 & 61.9 & 68.0 & \underline{52.5} & 55.9 & 49.9 & \underline{41.8} & 56.1 & 46.9 & 53.3\\
    Cai~\etal~\cite{cai2019exploiting}& ICCV'19 & 46.5 & 48.8 & 47.6 & 50.9 & 52.9 & 61.3 & 48.3 & \underline{45.8} & 59.2 & 64.4 & 51.2 & 48.4 & 53.5 & 39.2 & 41.2 & 50.6\\
    Zeng~\etal~\cite{zeng2020srnet}& ECCV'20 & - & - & - & - & - & - & - & - & - & - & - & - & - & - & - & 49.9\\
    Lin~\etal~\cite{lin2021end}& CVPR'21 & - & - & - & - & - & - & - & - & - & - & - & - & - & - & - & 54.0\\
    Li~\etal~\cite{li2021human}& ICCV'21 & - & - & - & - & - & - & - & - & - & - & - & - & - & - & - & 48.6\\
    Wehrbein~\etal~\cite{wehrbein2021probabilistic}& ICCV'21 &\underline{38.5} & \underline{42.6} & \textbf{39.9} & \textbf{41.7} & \underline{46.5} & 51.6 & \underline{39.9} & \textbf{40.8} & \underline{49.5} & 56.8 & \underline{45.3} & \underline{46.4} & 46.8 & \underline{37.8} & \underline{40.4} & \underline{44.3}\\
    \hline
    \textbf{WSP (Ours)}&  & \textbf{38.2} & \textbf{42.5} & \underline{40.5} & \underline{42.1} & \textbf{45.3} & \underline{46.6} & \textbf{34.5} & 46.2 & \textbf{49.1} & 62.4 & \textbf{44.2} & \textbf{39.6} & 43.1 & \textbf{34.8} & \textbf{39.7} & \textbf{43.2}\\
    \hline
    PA MPJPE(mm) & & Dir. & Dis. & Eat. & Gre. & Phoe. & Pho. & Pos. & Pur. & Sit. & SiD. & Smo. & Wai. & WaD. & Wal. & WaT. & Avg\\
    \hline
    Moreno~\etal~\cite{moreno20173d}& CVPR'17 & 67.4 & 63.8 & 87.2 & 73.9 & 71.5 & 69.9 & 65.1 & 71.7 & 98.6 & 81.3 & 93.3 & 74.6 & 76.5 & 77.7 & 74.6 & 76.5\\
    Zhou~\etal~\cite{zhou2017towards}& TPAMI'18 & 47.9 & 48.8 & 52.7 & 55.0 & 56.8 & 49.0 & 45.5 & 60.8 & 81.1 & 53.7 & 65.5 & 51.6 & 50.4 & 54.8 & 55.9 & 55.3\\
    Martinez~\etal~\cite{martinez2017simple}& ICCV'17 & 39.5 & 43.2 & 46.4 & 47.0 & 51.0 & 41.4 & 40.6 & 56.5 & 69.4 & 49.2 & 56.0 & 45.0 & 38.0 & 49.5 & 43.1 & 47.7\\
    Sun~\etal~\cite{sun2017compositional}& ECCV'17 & 42.1 & 44.3 & 45.0 & 45.4 & 51.5 & 43.2 & 41.3 & 59.3 & 73.3 & 51.0 & 53.0 & 44.0 & 38.3 & 48.0 & 44.8 & 48.3\\
    Fang~\etal~\cite{fang2018learning}& AAAI'18 & 38.2 & 41.7 & 43.7 & 44.9 & 48.5 & 40.2 & 38.2 & 54.5 & 64.4 & 47.2 & 55.3 & 44.3 & 36.7 & 47.3 & 41.7 & 45.7\\
    Sun~\etal~\cite{sun2018integral}& ECCV'18 & 36.9 & 36.2 & 40.6 & 40.4 & 41.9 & 34.9 & 35.7 & 50.1 & 59.4 & 40.4 & 44.9 & 39.0 & 30.8 & 39.8 & 36.7 & 40.6\\
    Cai~\etal~\cite{cai2019exploiting}& ICCV'19 & 36.8 & 38.7 & 38.2 & 41.7 & 40.7 & 46.8 & 37.9 & 35.6 & 47.6 & 51.7 & 41.3 & 36.8 & 42.7 & 31.0 & 34.7 & 40.2\\
    Moon~\etal~\cite{moon2019camera} & ICCV'19 & 31.0 & \underline{30.6} & 39.9 & 35.5 & \underline{34.8} & \textbf{30.2} & 32.1 & 35.0 & 43.8 & \textbf{35.7} & 37.6 & \underline{30.1} & \underline{24.6} & 35.7 & 29.3 & 34.0\\
    Wehrbein~\etal~\cite{wehrbein2021probabilistic}& ICCV'21 &\underline{27.9} & 31.4 & \underline{29.7} & 30.2 & 34.9 & 37.1 & \underline{27.3} & \textbf{28.2} & \underline{39.0} & 46.1 & \underline{34.2} & 32.3 & 33.6 & \underline{26.1} & \underline{27.5} & \underline{32.4}\\

    \hline
    \textbf{WSP (Ours)}&  & \textbf{25.1} & \textbf{27.7} & \textbf{26.1} & \textbf{27.2} & \textbf{31.3} & \underline{32.1} & \textbf{25.6} & \underline{30.6} & \textbf{31.8} & \underline{37.5} & \textbf{28.1} &\textbf{24.8}& \textbf{24.2} & \textbf{25.6} & \textbf{26.1} & \textbf{28.2} \\
    \hline
    
    \hline
  \end{tabular}
 }
 \label{table:hm36_p1}
\end{table}

\subsection{Ablation Study}
\subsubsection{Relative Depth Learning}
We formulate the pre-training task as a classification task, which aims to learn to distinguish the relative depth relationship between two human joints. After training WSP on the MCPC dataset, we test WSP on the test set of the MCPC dataset. The results of classification are shown in Table \ref{table:rdl_scale_eval} and Table \ref{table:rdl_crop_eval}. 

In Table \ref{table:rdl_scale_eval}, WSP-A, WSP-B, WSP-C, and WSP-D represent different test settings. All of the settings are WSP with the ResNet-50 backbone. The test images are cropped from the images in the testing set of MCPC. The cropped image includes two persons, which are the targets to distinguish relative depth relationships. Since the scales of the two human bodies in the image may be very different, we define the scale gap ratio $\Delta S$ to represent this scale gap. WSP-A, $\Delta S \leq 1.0$ means that all the cropped images are used for testing, regardless of the huge scale gap between the two persons in this cropped image. WSP-B, $\Delta S \leq 0.5$ means that only a part of cropped images is used for testing. The scale gap of two persons in the cropped image must be less than or equal to 0.5, which means the testing images are more difficult to distinguish relative depth relationships. With the same knowledge, WSP-C ($\Delta S \leq 0.3$) and WSP-D ($\Delta S \leq 0.1$) mean the smaller and more difficult testing set. 

WSP-A obtains 95.39\% in AUC. The values of accuracy, precision, recall, and F1 score are all over 88\%. These results illustrate that WSP can distinguish the relative depth relationship of keypoints when given an image with two persons. 
WSP-B, WSP-C, and WSP-D are designed to study the influence of the scale. The smaller $\Delta S$ means that the two persons in testing images are with more similar scales. As shown in the results in Table \ref{table:rdl_scale_eval}, all the evaluation metrics decrease with the smaller scale gap ratio $\Delta S$, which verifies that scale is an important visual cue to distinguish relative depth.
However, although we just keep the hardest cases for evaluation in WSP-D, we still can obtain over 90\% in AUC and over 81\% in accuracy, precision, recall, and F1 score, respectively. The results show that WSP can estimate relative depths successfully when two persons have almost the same scales.

Due to the relative depth supervision being generated according to the ankle cues, we study the influence of ankle cues in Table \ref{table:rdl_crop_eval}. RA in Table \ref{table:rdl_crop_eval} means removing the ankle parts in the images for testing. WSP-E, WSP-F, WSP-G, and WSP-H are designed to that WSP still can estimate relative depths successfully without the guiding of ankle clues.
Although there are no ankle joints in the input images, for the testing images with almost the same scales, WSP-H still can obtain over 87\% in AUC and over 79\% in accuracy, precision, recall, and F1 score, respectively. The results show that the accuracy of estimating relative depths is influenced by ankle joints, but WSP can still have a good estimation of relative depths even without the clues of ankle joints.

\subsubsection{3D Human Pose Estimation based on WSP}
\label{sec:3dphm36}
To verify that WSP is beneficial to 3D human pose estimation, we finetune WSP on the Human3.6M dataset.
The baseline is PoseNet~\cite{sun2018integral,kocabas2019self,moon2019camera}, which is a widely-used network for 3D human pose estimation. PoseNet and WSP are the same networks, the difference is that WSP is pre-trained on the MCPC dataset.
As shown in Table \ref{table:finetune_hm36}, PoseNet achieves 67.5 in MPJPE of action average based on ResNet-50. WSP achieves 54.3 in MPJPE and a relative gain of 20\% with the pre-trained weights.

In order to further analyze the influence of pre-training weights, we make a comparison of PoseNet and WSP on MPJPE of joints average. We can find that the MPJPE of joints averages on $Z$ coordinates are obviously greater than the MPJPE of joints average on $X$ and $Y$ coordinates from Table \ref{table:finetune_hm36}. Here, we denote them as $MPJPE_X$, $MPJPE_Y$, and $MPJPE_Z$, respectively. $MPJPE_Z$ of PoseNet are greater than 50mm, while $MPJPE_X$ and $MPJPE_Y$ of PoseNet are about 23mm. The phenomenon shows that the error of depth predicting is the key problem to improve the performance of 3D human pose estimation.
However, WSP obtains 40.2 in $MPJPE_Z$ of action average based on ResNet-50. Compared with PoseNet, the $MPJPE_Z$ decay to 40mm from 50mm, which verifies that the pre-trained WSP is beneficial to 3D pose estimation, especially in the depth prediction. We believe that the relative depth pre-training on MCPC learns how to distinguish the relative depth, which improves 3D pose estimation.

\subsubsection{MTS vs WSP}
Since we use the MCPC dataset for the pre-training of WSP, to verify the improvements in Section \ref{sec:3dphm36} come from more data or our pre-training strategy, we make the comparison of mixing training strategy (MTS) and WSP in this section.

MTS is adding 2D datasets with 3D datasets for mixing training. MTS just gives 2D supervision when using images in 2D datasets while 3D supervision when using images in 3D datasets. Following \cite{sun2018integral,kocabas2019self,moon2019camera}, we also finetune WSP on Human3.6M and MCPC datasets to ensure that WSP and MTS are trained on the same data for a fair comparison. The results are shown in Table \ref{table:finetune_hm36}. PoseNet achieves 52.9 MPJPE based on ResNet-50. WSP achieves 48.7 in MPJPE based on ResNet-50. Compared with PoseNet, the relative improvement is 6\%. The $MPJPE_Z$ of WSP decays to 35.5mm with ResNet-50, which has an improvement of 11\% in estimating depths compared with 39.9mm of PoseNet. The MTS tries to improve the performance of predicting 3D pose by adding the constraint of the 2D pose, which to further enhance the structural constraint of the 3D pose. However, WSP tries to directly learn weak 3D representation from 2D datasets, which significantly improves the accuracy of depth prediction for 3D human pose estimation as the results are shown in Table \ref{table:finetune_hm36}.

Therefore, the WSP improves the performance of 3D pose estimation since the relative depth pre-training. 
\subsection{Comparison with State-of-the-art Methods}
\subsubsection{Human3.6M Dataset}
We compare our proposed system with the state-of-the-art 3D human pose estimation methods on the Human3.6M dataset. Following the previous works~\cite{moon2019camera}, WSP is finetuned on Human3.6M dataset based ResNet-50. For evaluation on \textit{\textbf{Subject Protocol 1}}, as shown in Table \ref{table:hm36_p1}, WSP obtains an action average MPJPE of 43.2mm, which is the state-of-the-art result for fair comparison. For evaluation on \textit{\textbf{Subject Protocol 2}}, as shown in Table \ref{table:hm36_p1}, WSP obtains an action average PA MPJPE of 28.2mm, which is the state-of-the-art result for fair comparison.
\begin{table}
  \small
  \centering
\renewcommand\tabcolsep{25pt}
  \caption{Comparison with state-of-the-art methods on multi-person MuPoTS-3D dataset.}
  \vspace{0.3cm}
  \resizebox{\textwidth}{!}{
  \begin{tabular}{lcccc}
    \hline
    
    \hline
    3DPCK(\%, 150mm) & Backbone & Category & All & Matched \\
    \hline
    Lcr-net~\cite{rogez2017lcr} & VGG16 & \multirow{5}{*}{Top-down} & 53.8 & 62.4 \\
    Lcr-net++~\cite{rogez2019lcr} & ResNet50 &  & 70.6 & 74.0 \\
    HG-RCNN~\cite{dabral2019multi} & ResNet50 &  & 71.3 & 74.2 \\
    HMOR~\cite{wang2020hmor} & ResNet50 &  & - & 82.0 \\
    PoseNet~\cite{moon2019camera} & ResNet50 &  & 81.8 & 82.5 \\
    \hline
    ORPM~\cite{mehta2018single} & ResNet50 & \multirow{3}{*}{Bottom-up} & 65.0 & 69.8 \\
    Xnect~\cite{mehta2019xnect} & SelecSLSNet &  & 70.4 & 75.8 \\
    SMAP~\cite{zhen2020smap} & Hourglass &  & 73.5 & 80.5 \\
    \hline
    \textbf{WSP (Ours)} & ResNet50 & Top-down & \textbf{82.4} & \textbf{83.2} \\
    
    \hline
    
    \hline
  \end{tabular}
  }
  \label{tab:MuCo}
\end{table}

\begin{table}
    \centering
  \renewcommand\tabcolsep{20pt}
  \caption{Comparsion with baseline approach based on Resnet-18 on MuPoTS-3D dataset with multiple thresholds. Smaller threshold means stricter evaluation metric.}
\vspace{0.2cm}
\resizebox{\textwidth}{!}{
  \begin{tabular}{c|c|cccccc}
    \hline
    
    \hline
    \multirow{2}{*}{Methods} &\multirow{2}{*}{Eval}& \multicolumn{6}{c}{3DPCK Thresholds} \\
    \cline{3-8}
    & & 100mm & 110mm & 120mm & 130mm & 140mm & 150mm\\
    \hline
    PoseNet~\cite{moon2019camera} & \multirow{3}{*}{All} & 57.6 & 63.1 & 68.1 & 70.4 & 76.0 & 79.0\\
    \textbf{WSP(Ours)} & &\textbf{59.5} & \textbf{65.1} & \textbf{69.9} & \textbf{71.9} & \textbf{77.2} & \textbf{80.1}\\
    $\Delta$ &  & $\uparrow$ 3.3\% & $\uparrow$ 3.2\% & $\uparrow$ 2.6\% & $\uparrow$ 2.1\% & $\uparrow$ 1.5\% & $\uparrow$ 1.4\%\\
    \hline
    PoseNet~\cite{moon2019camera} & \multirow{3}{*}{Matched} & 58.6 & 64.2 & 69.4 & 73.8 & 77.5 & 80.5\\
    \textbf{WSP(Ours)} & & \textbf{60.2} & \textbf{65.9} & \textbf{70.7} & \textbf{74.7} & \textbf{78.2} & \textbf{81.2}\\
    $\Delta$ &  & $\uparrow$ 2.7\% & $\uparrow$ 2.6\% & $\uparrow$ 1.8\% & $\uparrow$ 1.2\% & $\uparrow$ 1.0\% & $\uparrow$ 0.9\%\\
    \hline
    
    \hline
  \end{tabular}
  }
  \label{table:muco_thres}
\end{table}

\subsubsection{MuPoTS-3D Datasets}
We compare our approach with state-of-the-art 3D human pose estimation methods on MuPoTS-3D Datasets in Table \ref{tab:MuCo}. MuCo-3DHP and MuPoTS-3D Datasets are outdoor multi-person 3D human pose estimation dataset. The evaluation metric is a 3D percentage of correct keypoints. Due to the multiple persons, the accuracy is computed for all ground-truth keypoints or ground-truth keypoints in matched persons. As shown in Table \ref{tab:MuCo}, WSP outperforms state-of-the-art methods in 3DPCK with a threshold of 150mm. 

To further study the effectiveness of WSP, we compare the results of WSP and PoseNet \cite{moon2019camera} on MuPoTS-3D datasets with more strict thresholds (100mm, 110mm, 120mm, 130mm, 140mm, and 150mm) since 150mm is a broad threshold. The results are shown in Table \ref{table:muco_thres}. The testing threshold decreases to 100mm from 150mm, and the 3DPCK of PoseNet and WSP decreases. However, for all-person test settings, compared with PoseNet, the relative improvement of WSP increases to 3.3\% from 1.4\% with the more strict thresholds. For matched-person settings, the relative improvement of WSP increases to 2.7\% from 0.9\% with the more strict thresholds.

\begin{figure}
    \centering
    \includegraphics[width=\textwidth]{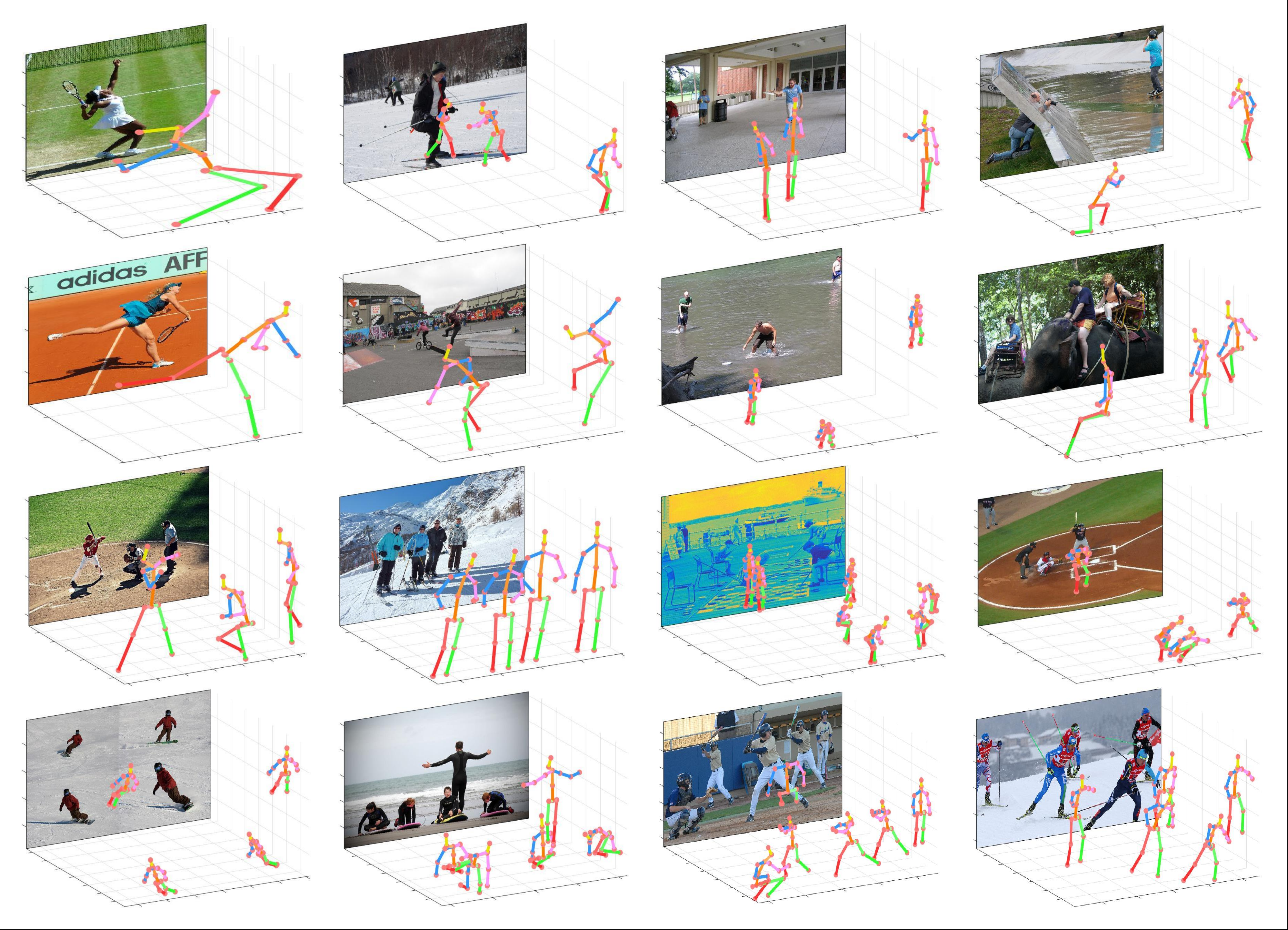}
    \caption{The qualitative results of WSP with ResNet-50 backbone on COCO dataset. We train WSP on the MuCo dataset and test WSP on the COCO dataset. Note that there is no 3D annotation on the COCO dataset. WSP performs well even on these in-the-wild images with huge pose variances.}
    \label{fig:f4}
\end{figure}

\begin{table*}
  \caption{Cross-action Evaluation on Human3.6M Dataset. Each entry shows that the model is trained only on the corresponding single action and tested on all actions.}
  \vspace{0.2cm}
  \renewcommand\tabcolsep{3pt}
  \resizebox{\textwidth}{!}{
  \begin{tabular}{lcccccccccccccccc}
    \hline
    
    \hline
   PA MPJPE(mm) & Dir. & Dis. & Eat. & Gre. & Phoe. & Pho. & Pos. & Pur. & Sit. & SiD. & Smo. & Wai. & WaD. & Wal. & WaT. & Avg\\
    \hline
    Martinez~\etal~\cite{martinez2017simple} & 127.3 & 104.6 & 95.0 & 116.1 & 95.5 & 117.4 & 111.4 & 125.0 & 116.9 & 93.6 & 111.0 & 126.0 & 131.0 & 106.3 & 140.5 & 114.5\\
    Ci~\etal~\cite{ci2019optimizing} & 102.8 & 88.6 & 79.1 & 99.3 & 80.0 & 91.5 & 93.2 & 89.6 & 90.4 & 76.6 & 89.6 & 102.1 & 108.8 & 90.8 & 118.9 & 93.4\\
    Zeng~\etal~\cite{zeng2020srnet} & 92.3 & 71.4 & 71.8 & 86.4 & 66.8 & 79.1 & 82.5 & 86.6 & 88.9 & 93.4 & \textbf{66.1} & 83.0 & \textbf{74.4} & 90.0 & 97.8 & 82.0\\
    \hline
    \textbf{WSP (Ours)} & \textbf{84.8} & \textbf{70.3} & \textbf{64.4} & \textbf{78.6} & \textbf{60.4} & \textbf{72.5} & \textbf{76.9} & \textbf{68.7} & \textbf{75.8} & \textbf{60.0} & 77.0 & \textbf{81.7} & 82.2 & \textbf{73.1} & \textbf{93.0} &\textbf{74.6}\\
    \hline
    
    \hline
  \end{tabular}
 }
 \label{table:cross_action}
\end{table*}

\subsection{Evaluation of Generalization}
The experimental results on Human3.6M~\cite{ionescu2013human3} and MuPoTS-3D~\cite{mehta2017monocular} datasets have shown the effectiveness of WSP. The muPoTS-3D dataset is collected in outdoor scenes. Thus, the improvements on the MuPoTS-3D dataset of WSP demonstrate the generalization ability of the WSP model.

To further verify the generalization ability of WSP, we make the comparison with previous works \cite{martinez2017simple,ci2019optimizing,zeng2020srnet} following the settings of \textit{\textbf{Cross Action Protocol}}. The results of cross-action evaluation are shown in Table \ref{table:cross_action}. All the results in Table \ref{table:cross_action} are obtained by WSP, which is just trained on the single-action images and tested on the all-action images. WSP yields an overall improvement of 7.4mm (relative 9\% improvement) over \cite{zeng2020srnet}. The results of the cross-action evaluation indicate the strong generalization ability of WSP obtained by the relative depth pre-training strategy.

In addition, we also predict 3D pose on COCO~\cite{lin2014microsoft} dataset, which is a large-scale 2D pose dataset in the wild scenes. Due to no 3D annotations on the COCO dataset, we train WSP on the MuCo dataset and test WSP on the COCO dataset. The qualitative results of applying our methods to the in-the-wild images are shown in Figure \ref{fig:f4}. The results of these in-the-wild images show that the WSP model has strong generalization ability.

\section{Conclusion}
To solve the problems of lacking enough 3D pose annotations and generalization, we propose a novel approach, named weakly-supervised pre-training (WSP), which extracts weak 3D information from monocular images without the supervision of 3D annotations. WSP is pre-trained on the 2D pose datasets and finetuned on the 3D pose datasets. For pre-training, we propose a large-scale 2D dataset (MCPC), with the labels of relative depth generated according to perspective knowledge. After finetuning WSP on two 3D pose datasets, WSP achieves state-of-the-art results. The results on the COCO dataset show the strong generalization ability of WSP. To the best of our knowledge, this is the first weakly-supervised pre-training approach for 3D pose estimation, which extracts weak 3D information from 2D pose datasets without manual 3D annotations.



 \bibliographystyle{elsarticle-num} 
 \bibliography{ref.bib}






\clearpage

\end{document}